\documentclass{article}

\usepackage{arxiv}
\usepackage{amsmath,amsfonts}
\usepackage[utf8]{inputenc} 
\usepackage[T1]{fontenc}    
\usepackage{hyperref}       
\usepackage{url}            
\usepackage{booktabs}       
\usepackage{amsfonts}       
\usepackage{graphicx}
\usepackage{doi}
\usepackage{amssymb}

\usepackage{makecell}
\usepackage{threeparttable}

\title{Stable Time Series Prediction of Enterprise Carbon Emissions Based on Causal Inference}

\newif\ifuniqueAffiliation

\uniqueAffiliationtrue

\ifuniqueAffiliation
\usepackage{authblk}

\newbox{\orcid}\sbox{\orcid}{\includegraphics[scale=0.06]{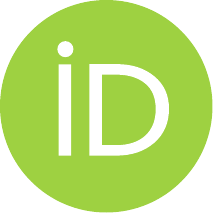}}
\author[1,2]{{\usebox{\orcid}\hspace{1mm}Zitao Hong}%
}
\author[1,2]{{\usebox{\orcid}\hspace{1mm}Zhen Peng\thanks{\texttt{Corresponding author:pengzhen@cugb.edu.cn}}}%
}
\author[1,2]{{\usebox{\orcid}\hspace{1mm}Xueping Liu}%
}
\affil[1]{School of Economics and Management, China University of Geosciences (Beijing), Beijing 100083, China}
\affil[2]{MOE Laboratory of Philosophy and Social Sciences for Mineral Resources Security Governance, China University of Geosciences (Beijing), Beijing 100083, China}

\begin{document}
\maketitle

\begin{abstract}
Against the backdrop of ongoing carbon peaking and carbon neutrality goals, accurate prediction of enterprise carbon emission trends constitutes an essential foundation for energy structure optimization and low-carbon transformation decision-making. Nevertheless, significant heterogeneity persists across regions, industries and individual enterprises regarding energy structure, production scale, policy intensity and governance efficacy, resulting in pronounced distribution shifts and non-stationarity in carbon emission data across both temporal and spatial dimensions. Such cross-regional and cross-enterprise data drift not only compromises the accuracy of carbon emission reporting but substantially undermines the guidance value of predictive models for production planning and carbon quota trading decisions. To address this critical challenge, we integrate causal inference perspectives with stable learning methodologies and time-series modelling, proposing a stable temporal prediction mechanism tailored to distribution shift environments. This mechanism incorporates enterprise-level energy inputs, capital investment, labour deployment, carbon pricing, governmental interventions and policy implementation intensity, constructing a risk consistency-constrained stable learning framework that extracts causal stable features—robust against external perturbations yet demonstrating long-term stable effects on carbon dioxide emissions—from multi-environment samples across diverse policies, regions and industrial sectors. Furthermore, through adaptive normalization and sample reweighting strategies, the approach dynamically rectifies temporal non-stationarity induced by economic fluctuations and policy transitions, ultimately enhancing model generalization capability and explainability in complex environments. Empirical analyses based on time-series data encompassing energy economics and carbon emissions from multiple national enterprises demonstrate that this method achieves superior robustness and accuracy in cross-regional and cross-industry prediction tasks, yielding reductions in average prediction error of approximately 10\% to 15\% compared with conventional models. Our findings reveal the stable driving mechanisms underlying enterprise carbon emissions, furnishing novel methodological insights and quantitative foundations for reliable carbon emission forecasting and enterprise production and carbon quota trading decisions.
\end{abstract}


\section{Introduction}
\subsection{Problem Background}
To address the challenges of cross-environmental data distribution disparities and non-stationary process modelling in enterprise carbon emission prediction, this study establishes a stable temporal prediction framework for carbon emissions under complex operational condition shifts. Carbon emission processes constitute a class of typical non-stationary industrial processes whose temporal evolution is driven not only by long-term variables such as production structure and energy inputs but also exhibits high sensitivity to external perturbations including macroeconomic fluctuations and policy adjustments, resulting in pronounced spatiotemporal distribution heterogeneity \cite{1}. The Intergovernmental Panel on Climate Change (IPCC) indicates that, absent effective enterprise-level carbon emission monitoring and prediction mechanisms within the coming decades, the global climate system will face irreversible risks \cite{2}.

In contrast to traditional regional carbon emission accounting, enterprise carbon emission processes are deeply embedded within production decisions, energy input structures and technological choices, characterized by more substantial micro-level heterogeneity and dynamic complexity \cite{3}. Existing studies suggest that, even within the same industrial sector, disparities in energy utilization efficiency, technological pathways and policy responsiveness across different enterprises may yield significant divergence in carbon emission evolution trajectories \cite{4}. Particularly under policy constraints such as carbon emission trading schemes and energy consumption dual-control regulations, enterprise carbon emission behaviours are undergoing a profound transformation from factor-driven to constraint-incentive dual-driven patterns, rendering their temporal dynamic characteristics increasingly intricate.

From a data characteristics perspective, enterprise carbon emission time series exhibit pronounced non-stationarity; furthermore, due to systematic disparities in resource endowments, industrial structures and policy implementation intensity across regions, the data demonstrate significant distribution heterogeneity across spatial dimensions \cite{5}. The superposition of temporal non-stationarity and spatial heterogeneity yields a typical cross-environmental operational condition shift problem for carbon emission prediction. Nevertheless, existing prediction models are predominantly trained on single-region or single-period data, implicitly assuming that data generating mechanisms remain relatively stable out-of-sample. When models are directly applied to new regions, industries or policy scenarios, predictive performance often degrades substantially, yielding systematic biases \cite{6,7}. This not only undermines the guidance value of carbon emission prediction for enterprise production planning and carbon asset management but also constrains the evaluation precision of data-driven carbon governance policies to a considerable extent.

Thus, under realistic conditions where distribution shifts and non-stationary environments are ubiquitous, how to construct enterprise carbon emission prediction models possessing cross-environmental stable generalization capabilities, thereby furnishing reliable support for enterprise-level low-carbon decision-making, has emerged as a critical problem urgently requiring resolution in the field of energy systems engineering.
\subsection{Existing Research Progress and Limitations}
Regarding non-stationary process prediction problems, existing research broadly falls into three categories: methods grounded in traditional econometrics, approaches based on machine learning and deep learning, and causal inference and stable learning methods that have gradually emerged in recent years.

Within the traditional econometric framework, researchers predominantly employ regression analysis, structural decomposition analysis (SDA), error correction models or ARIMA-class time series models to characterize long-term evolution trends and driving factors of carbon emissions \cite{8}. Such methods emphasize long-term equilibrium relationships among variables, offering certain advantages in explaining structural linkages among economic growth, energy consumption and carbon emissions. Nevertheless, they are typically constrained by assumptions of parameter stability and stationarity, exhibiting limited adaptability to sudden shocks, institutional policy transitions and structural breakpoints \cite{9}. When data generating mechanisms undergo changes, model parameters frequently require re-estimation, rendering predictive stability difficult to guarantee.

With improvements in computational capabilities and data availability, machine learning and deep learning methods have been extensively introduced into carbon emission prediction research, including support vector regression (SVR), random forest (RF), long short-term memory (LSTM) networks and their variants \cite{10}. Related studies demonstrate that, for in-sample or short-term prediction tasks, such methods typically achieve substantial predictive accuracy. Nevertheless, they remain intrinsically data-driven approaches grounded in the empirical risk minimization (ERM) framework, exhibiting high sensitivity to training data distributions. When application environments shift, models readily exploit spurious correlations that hold only under specific scenarios, resulting in significant out-of-sample performance degradation \cite{11}.

Notably, whether traditional econometric methods or machine learning approaches, existing research predominantly adopts prediction accuracy as the core evaluation criterion, failing to adequately distinguish between the causal versus correlational roles of different explanatory variables. This problem becomes particularly pronounced under conditions of frequent policy adjustments or drastic economic fluctuations: certain variables may demonstrate high correlation with carbon emissions during specific periods, yet their effects lack structural stability, causing predictive relationships to rapidly collapse once environmental conditions change \cite{12,13}.

Against this backdrop, causal inference and stable learning theory furnish novel research perspectives for addressing distribution shift problems. Stable learning emphasizes identifying causal relationships that remain invariant across different environments through multi-environment data, thereby achieving robust prediction for unknown environments \cite{14}. In recent years, related methods have achieved progress in medical diagnosis, survival analysis and domain generalization problems \cite{15}. Nevertheless, existing stable learning research has predominantly focused on static classification tasks, with systematic applications in time series prediction and energy economics remaining relatively limited. Research specifically targeting enterprise- level carbon emissions (a typical non-stationary industrial process) remains in its nascent stages.
\subsection{Research Motivation}
To circumvent the limitations of aforementioned methods regarding cross-environmental generalization capability and causal stability, this study re-examines enterprise carbon emission prediction problems from the perspective of non-stationary industrial process modelling. The fundamental challenge lies in disentangling two distinct categories of influencing factors: one comprises long-term structural causal features determined by energy structure and production factor allocation, whose effects on carbon emissions possess relatively stable causal interpretations; the other encompasses short-term perturbations induced by business cycle fluctuations, energy price shocks or policy adjustments, whose effects typically exhibit phase-dependent and environment-specific characteristics \cite{5}. Unless effectively distinguished within model structures, predictive outcomes will inevitably compromise stability in cross-environmental applications.

Furthermore, although stable learning methods theoretically possess the potential to identify causally stable features, how they can be systematically integrated with temporal dynamic modelling and how dynamic effects of economic fluctuations on predictive relationships can be characterized remain underexplored. Consequently, it is necessary to systematically introduce stable learning concepts into temporal prediction modelling for enterprise carbon emissions, establishing a synergistic mechanism between causal invariance and temporal dynamic characteristics.
\subsection{Main Contributions}
To address spurious causal interference and inadequate model generalization capability faced by enterprise carbon emission prediction in non-stationary environments, this study proposes a cross-environmental carbon emission temporal prediction method grounded in stable learning. This method leverages stable learning to identify and optimize causal consistency features across different data distributions, ensuring that extracted causal drivers remain stable across cross-regional and cross-industry environments. On this foundation, we incorporate adaptive normalization and sample reweighting mechanisms to capture dynamic dependencies in temporal data, characterizing spatiotemporal evolution properties of causal features. Additionally, we adopt sample reweighting techniques based on covariate balancing, adjusting sample weights to enable models to more accurately reflect causal characteristics under ideal conditions.

The innovations and contributions of this study are threefold:
\begin{itemize}
  \item To address cross-environmental prediction challenges for non-stationary industrial processes, we propose an enterprise carbon emission prediction framework integrating stable learning with dynamic temporal modelling, establishing theoretical linkages between predictive stability and causal invariance;
  \item To address spurious causal relationship identification problems, we employ sample reweighting decorrelation operators from stable learning to eliminate spurious correlations induced by confounding factors among variables, extracting causal features insensitive to distribution shifts;
  \item We develop a stable learning-driven enterprise carbon emission temporal prediction model (Stable-CarbonNet), validating its predictive stability and generalization performance in cross-regional and cross-industry scenarios through multi-environment panel data, thereby furnishing reliable model support for enterprise low-carbon decision-making.
\end{itemize}
\section{Related Work}\label{RW}
\subsection{Traditional Econometric Methods and Their Limitations}
To address problems of long-term trend characterization and driving mechanism analysis for enterprise carbon emissions, traditional econometric methods constituted the dominant research paradigm in early-stage investigations of this field. Related studies typically employed regression analysis, structural decomposition analysis (SDA) and time series models to characterize carbon emission trends and their underlying driving mechanisms \cite{9}. Such approaches emphasize long-term equilibrium relationships among variables, particularly focusing on stable linkages among economic growth, energy consumption structure and carbon emissions, yielding substantial research outcomes at both macroscopic and microscopic levels \cite{10,16}.

In enterprise or regional-level research, regression models have been extensively utilized to identify key factors influencing carbon emission levels, including energy input structures, capital intensity, technological progress and industrial structure changes \cite{17}. Certain studies further incorporate fixed effects or dynamic panel models to control for unobservable individual heterogeneity, thereby enhancing the robustness of estimation results \cite{18}. Meanwhile, SDA methods furnish a clear analytical framework for understanding long-term carbon emission evolution mechanisms by decomposing emission changes into scale effects, structural effects and technological effects \cite{19}.

Nevertheless, as carbon emission governance enters a policy-intensive period, the inherent limitations of these approaches have become increasingly apparent. On the one hand, traditional econometric models typically implicitly assume parameter stability and data stationarity, whereas enterprise carbon emission data in practice are frequently subjected to policy adjustments, energy price fluctuations and macroeconomic shocks, resulting in pronounced structural break characteristics \cite{20}. Existing studies demonstrate that, under conditions involving structural breakpoints or institutional transitions, predictive performance of traditional time series models degrades substantially \cite{16}. On the other hand, such methods rely predominantly on ex post interpretation, exhibiting limited extrapolation capability for future unknown policy environments \cite{22}.

In recent years, scholars have begun to address econometric modelling problems under non-stationary environments, for instance by introducing structural break tests, piecewise regression or time-varying parameter models to characterize dynamic evolution of carbon emission relationships \cite{23}. Although these methods alleviate parameter instability to a certain extent, they remain primarily confined to descriptive correction levels, failing to fundamentally guarantee predictive stability across environmental scenarios at the mechanistic level. Consequently, for enterprise carbon emission prediction problems, sole reliance on traditional econometric frameworks proves inadequate for addressing complex real-world demands.
\subsection{Machine Learning Methods and Their Generalization Bottlenecks}
With advancements in big data technologies and computational capabilities, machine learning and deep learning methods have been progressively introduced into carbon emission prediction research, demonstrating substantial advantages in short-term prediction accuracy \cite{24}. Related studies indicate that models based on support vector regression (SVR), random forest (RF), gradient boosting decision trees (GBDT) and neural networks can effectively capture nonlinear characteristics in carbon emission data, thereby achieving lower prediction errors for in-sample or short-term prediction tasks \cite{25}.

In the deep learning direction, long short-term memory (LSTM) networks, gated recurrent units (GRU) and their variants have been widely adopted to characterize long-term dependency structures in carbon emission time series \cite{26}. Certain studies further enhance model fitting capabilities for complex dynamic patterns by introducing attention mechanisms, multi-scale feature extraction or hybrid model structures \cite{27}. In energy economics research, such methods have been applied to carbon emission prediction at national, regional and enterprise levels, demonstrating superior predictive accuracy relative to traditional models in comparative experiments \cite{28}.

Nevertheless, substantial empirical studies demonstrate that machine learning and deep learning models exhibit high dependence on training data distribution characteristics, remaining intrinsically grounded in the empirical risk minimization (ERM) paradigm. When disparities exist between training and prediction environments, models readily exploit spurious correlations that hold only within specific samples rather than universal causal relationships \cite{29}. Under circumstances involving policy adjustments, drastic energy price fluctuations or industrial structure transformations, such correlation dependencies often yield rapid performance degradation \cite{30}.

In carbon emission prediction applications, existing studies have identified that deep models exhibit high fitting accuracy in-sample yet demonstrate significantly insufficient stability in cross-regional, cross-industry or cross-period prediction tasks \cite{31}. This indicates that sole reliance on complex model structures cannot fundamentally resolve distribution shift problems. How to enhance cross- environmental stability while maintaining predictive accuracy constitutes a critical challenge for further applications of machine learning methods in energy economics.
\subsection{Causal Inference and Stable Learning Theory}
To address the limitations of aforementioned methods regarding cross- environmental generalization capability and causal stability, causal inference and stable learning theory furnish novel theoretical perspectives for tackling distribution shift problems. The core insight lies in identifying causal driving factors that maintain stable relationships across different environments rather than relying on surface statistical correlations susceptible to failure. Traditional ERM methods exhibit a propensity to exploit correlation patterns most amenable to fitting during training; nevertheless, these patterns frequently lack cross-environmental stability \cite{16,32}, proving insufficient to support long-term reliable prediction.

Stable learning aims to extract stable causal features that remain invariant despite environmental changes from multi-environment data, thereby enhancing model performance in unknown distributions \cite{15,30}. Existing studies demonstrate that leveraging causal invariance principles for feature selection, sample reweighting or structural modelling can substantially enhance generalization capability across diverse tasks \cite{13}. Specifically, related research has begun to explore applications of causal concepts for prediction tasks: Bodendorf et al. \cite{33} combined deep models with structural equation models for supply chain prediction; Yousefi et al. \cite{34} achieved watershed hydrological prediction through causal empirical decomposition; Li et al. \cite{35} embedded causal inference into decomposition-prediction frameworks for carbon market price prediction. These approaches collectively illustrate a consistent scientific principle: only features possessing genuine causal effects can maintain predictive efficacy when environments change \cite{36,37}. Such causality-driven robustness analysis can furnish more reliable foundations for carbon emission prediction under complex environmental conditions \cite{38,39,40}.

Nevertheless, existing methods have predominantly targeted static tabular data or macroscopic time series, leaving a critical scientific question insufficiently addressed: how can stable causal features be extracted from enterprise-level non-stationary temporal data to achieve robust prediction across environments? At the application level, stable learning methods have been employed in medical prediction, survival analysis and domain generalization problems, achieving certain efficacy \cite{41,42}. Nevertheless, compared with static data or classification tasks, time series prediction confronts more complex non-stationarity and dynamic feedback problems. Existing studies indicate that how to characterize causal invariance along the temporal dimension and handle situations where causal effects vary over time remain formidable challenges in stable learning theory applications \cite{43}.

In the field of energy economics, although a few studies have attempted to introduce causal inference ideas into price forecasting or energy demand forecasting \cite{44}, existing methods exhibit significant methodological limitations. Specifically, literature \cite{44} focuses on the price formation mechanism of the electricity market, utilizing causal modeling and structural equation modeling to identify structural associations between market supply and demand, fuel prices, and electricity prices. However, this approach primarily concentrates on static equilibrium analysis at the macro-market level, implicitly assuming that the data generation process satisfies stationarity and parameter stability, without fully considering the heterogeneity of micro-enterprise behaviors and structural breaks caused by policy shocks. More importantly, the causal inference framework employed in literature \cite{44} lacks cross-environmental generalization ability: its model parameter estimation relies on sample distributions from specific market periods, making it difficult to ensure the stability of predictive relationships when facing policy system changes (such as the introduction of carbon trading mechanisms) or energy structure transformation. Furthermore, this method does not incorporate a stable learning mechanism, making it impossible to effectively distinguish between causally stable features and spurious correlations that only hold in specific environments, leading to significant degradation in the model's predictive performance when transferred across regions or policy stages. This current situation provides an important entry point for the research presented in this paper.
\subsection{Literature Assessment}
In summary, existing enterprise carbon emission prediction research has undergone a rough evolutionary trajectory from traditional econometric models to machine learning models, yet exhibits substantial deficiencies in addressing distribution shifts and non-stationary environments. On the one hand, traditional econometric methods offer advantages in explaining long-term trends yet demonstrate limited adaptability to structural changes; on the other hand, although machine learning methods achieve prominent performance in fitting accuracy, their predictive relationships lack causal stability, rendering reliable cross-environmental performance difficult to maintain.

From a theoretical standpoint, causal inference and stable learning furnish novel analytical frameworks for resolving these problems, yet their applications in temporal prediction and energy economics remain to be deepened. Particularly for enterprise carbon emission prediction, how to simultaneously identify long-term stable causal driving mechanisms while effectively correcting for short-term economic fluctuations and policy shocks has yet to coalesce into a mature, unified framework.

Motivated by these considerations, this study systematically introduces stable learning concepts into temporal prediction problems for enterprise carbon emissions from a causal inference perspective, aiming to redress deficiencies in existing research regarding causal stability and temporal non-stationarity. By constructing cross- environmental risk consistency constraints and integrating dynamic correction mechanisms, we endeavor to furnish novel research paradigms for enterprise carbon emission prediction under complex environmental conditions at both theoretical and methodological levels.

\section{Method}\label{22}
\subsection{Problem Formulation}
To address the distribution shift and temporal non-stationarity characteristics exhibited by enterprise carbon emissions across multi-policy and multi-regional environments, we propose a cross-environmental prediction framework integrating causal stable learning with temporal dynamic modelling, which we designate as Stable-CarbonNet. This framework leverages stable learning to identify causally stable features insensitive to diverse environments while introducing temporal adaptive mechanisms to capture dynamic variations, thereby ensuring predictive stability during cross-environmental migration.

Against the backdrop of concurrent multi-policy, multi-regional and multi- industry contexts, enterprise carbon emission behaviours are not generated by a singular stable mechanism but are rather driven by a constellation of economic structural factors evolving across temporal and spatial dimensions. Particularly during the ongoing advancement of dual-carbon objectives, factors including energy price fluctuations, industrial structure adjustments and regulatory intensity changes frequently introduce exogenous shocks, resulting in pronounced distribution shifts and temporal non-stationarity in enterprise carbon emission data. This characteristic determines that enterprise carbon emission prediction cannot be simplistically regarded as conventional time series forecasting tasks but rather approximates a cross-environmental non-stationary generalization problem.

To characterize this problem, let $e_t\in\varepsilon$ denote the macro-policy environment where enterprise $i$ operates at time $t$, with its carbon emission level denoted as $Y_{i,t}^{(e_t)}$ and the corresponding explanatory variable vector as $X_{i,t}^{(et)}\in\mathbb{R}^p$. Without discrimination, the carbon emission generation mechanism can be formalized as:
\begin{equation}\label{eq1}
Y_{i,t}^{(e_t)} = g^{(e)}(X_{i,t}) + \varepsilon_{i,t}^{(e_t)},
\end{equation}
where $g^{(e)}(\cdot)$ explicitly depends on environment $e$, indicating that the mechanisms through which enterprise characteristics influence carbon emissions differ across policy or macroeconomic conditions.

Should we directly model based on equation (\ref{eq1}), model parameters would inherently depend on specific environmental distributions, yielding substantial performance degradation during environment switching. The core insight of this study lies in the recognition that, despite the overall generation mechanism $g^{(e)}(\cdot)$ varying with environments, a portion of environmentally insensitive stable causal structures persists therein.
\subsection{Structural Decomposition from a Causal Stability Perspective}
Grounded in causal inference theory, the enterprise carbon emission generation process can be decomposed into the superposition of stable structural components and environmental perturbation terms:
\begin{equation}\label{eq2}
Y_{i,t}^{(e_t)} = f^*(X_{i,t}^c) + h^{(e)}(X_{i,t}^s) + \varepsilon_{i,t}^{(e_t)},
\end{equation}
where $X^c$ represents the core feature subset possessing cross-environmental causal stability regarding carbon emissions, such as long-term energy structures and capital equipment levels; $X^s$ denotes short-term or institutional variables merely correlated with carbon emissions under specific environments;$f^{*}(\cdot)$ indicates the environment-invariant structural function; and $h^{(e)}(\cdot)$ represents the environment-varying disturbance mechanism.

The crucial implication of equation (\ref{eq2}) lies in the following: provided that predictive models can accurately identify and rely upon $f^*(\cdot)$, stable performance can be maintained despite environmental changes. Nevertheless, as $X^c$ and $X^s$ are frequently highly mixed at the observation level, direct discrimination between the two proves infeasible during practical modelling, necessitating indirect identification through cross-environmental constraints.

According to cross-environmental risk consistency and stable learning principles, let the predictive model be parameterized as $f_\theta(\cdot)$, with its risk function under environment $e$ defined as:
\begin{equation}\label{eq3}
R_e(\theta) = \mathbb{E}_{(X,Y) \sim \mathcal{D}_e}[\mathcal{L}(f_\theta(X), Y)],
\end{equation}
where $\mathcal{L}(\cdot)$ denotes the squared or absolute loss function. Traditional empirical risk minimization methods are equivalent to solving:
\begin{equation}\label{eq4}
\min_\theta \sum_{e \in \mathcal{E}} R_e(\theta),
\end{equation}
yet this objective fails to constrain consistency of optimal parameters across different environments. The fundamental insight of stable learning lies in the following: truly causally stable predictive mechanisms should simultaneously optimize across all environments. To this end, we introduce a shared representation function $\Phi_\theta(\cdot)$ and linear prediction head $w$, reparameterizing the model as:
\begin{equation}\label{eq5}
f_\theta(X) = w^T \Phi_\theta(X).
\end{equation}

Upon this foundation, stability requirements can be formalized as:
\begin{equation}\label{eq6}
w \in \arg\min_\theta \sum_w R_e(w^T \Phi_\theta(X)), \forall e \in \mathcal{E}.
\end{equation}

As equation (\ref{eq6}) constitutes an implicit constraint, we employ gradient consistency as its first-order approximation. Should $w$ be optimal across all environments, the following must be satisfied:
\begin{equation}\label{eq7}
\nabla_w R_e(w^T \Phi_\theta(X))=0, \forall e \in \mathcal{E}.
\end{equation}

We further define the gradient discrepancy metric:
\begin{equation}\label{eq8}
S(\theta, w) = \sum_{e \in \mathcal{E}} \left\| \nabla_w R_e(w^T \Phi_\theta(X)) \right\|_2^2,
\end{equation}
which approaches zero, indicating consistent optimality judgments regarding predictive parameters across all environments. Integrating this stability constraint into the original risk minimization framework yields the joint optimization objective:
\begin{equation}\label{eq9}
\min_{\theta,w} \sum_{e \in \mathcal{E}} R_e(w^T \Phi_\theta(X)) + \lambda S(\theta, w),
\end{equation}
where $\lambda$ controls the trade-off between stability and fitting accuracy.

\subsection{Modelling Temporal Non-stationarity}

To address heteroscedasticity and structural break characteristics present in enterprise carbon emission time series data, upon the foundation of the stable learning framework presented in Section 3.2, we introduce a temporal weighting function $\omega_t$ to adaptively adjust the importance of samples from different periods:
\begin{equation}\label{eq10}
R_e(\theta) = \mathbb{E}\left[\omega_t \cdot \mathcal{L}(f_\theta(X_t), Y_t)\right],
\end{equation}
where the weighting function $\omega_t$ is determined jointly by historical volatility magnitude and policy change intensity:
\begin{equation}\label{eq11}
\omega_t = \exp(-\alpha \cdot \Delta_t),
\end{equation}
with $\Delta_t$ representing the degree of statistical distribution variation before and after time $t$. Furthermore, to mitigate interference from environmental scale differences on model learning, we incorporate an adaptive normalization operator $N_e(\cdot)$, mapping input features from diverse environments onto comparable scales:
\begin{equation}\label{eq12}
\tilde{X}_{i,t}^{(e)} = N_e(X_{i,t}^{(e)}).
\end{equation}

Integrating stability constraints with temporal correction mechanisms, the ultimate model optimization objective can be formulated as:
\begin{equation}\label{eq13}
\min_{\theta,w} \sum_{e \in \mathcal{E}} \mathbb{E}\left[\omega_t \cdot \mathcal{L}(w^T \Phi_\theta(N_e(X)), Y)\right] + \lambda \sum_{e \in \mathcal{E}} \left\| \nabla_w R_e(w^T \Phi_\theta(X)) \right\|_2^2.
\end{equation}

Taking gradients with respect to the parameters yields the update directions:
\begin{equation}\label{eq14}
\theta^{(k+1)} = \theta^{(k)} - \eta \frac{\partial \mathcal{L}}{\partial \theta},\; w^{(k+1)} = w^{(k)} - \eta \frac{\partial \mathcal{L}}{\partial w}.
\end{equation}

Under limiting conditions, when $\lambda \to \infty$, the model degenerates into a fully causally stable predictor; when $\lambda = 0$, the model degenerates into an ordinary ERM temporal prediction model:
\begin{equation}\label{eq15}
\lim_{\lambda \to 0} f_\theta = \text{ERM},\ \  \lim_{\lambda \to \infty} f_\theta = \text{Invariant Predictor}.
\end{equation}

\begin{figure}[ht]
\centering
  \includegraphics[width=4.5in, height=2.8in]{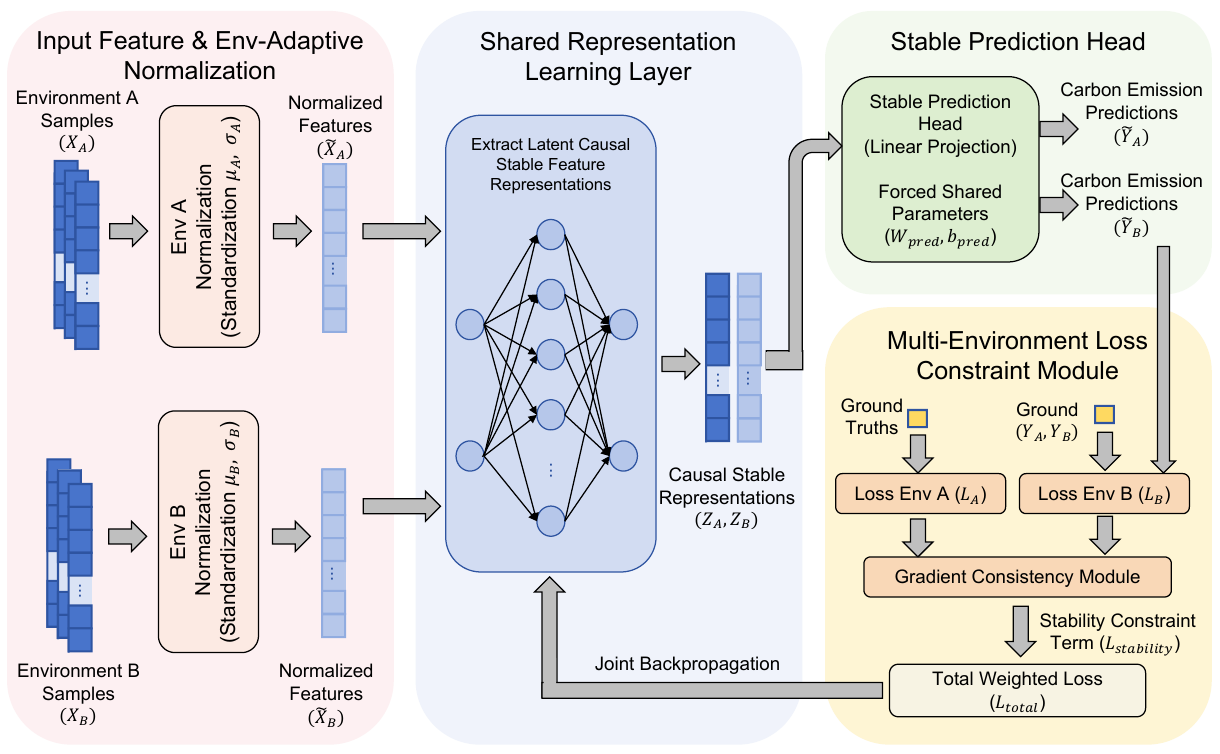}
    \renewcommand\baselinestretch{0.85}
    \vspace{-0.2cm}
\caption{Architecture of the proposed model.}\label{optimal_price_colume}
\end{figure}

The overall architecture of our proposed model is illustrated in Figure \ref{optimal_price_colume}, comprising five sequentially connected components: the input feature layer, environment-adaptive normalization layer, shared representation learning layer, stable prediction head and multi-environmental loss constraint module. Initially, enterprise samples from heterogeneous environments enter corresponding normalization sub-modules for intra-environmental standardization, achieving preliminary alignment of feature distributions without explicitly introducing environmental labels. Subsequently, processed features flow into fully parameter-shared representation learning networks, aiming to strip environment-specific disturbances and extract deep causally stable features. Thereafter, extracted feature vectors are mapped to carbon emission predictions via the linear prediction head, which maintains strong parameter consistency across all environments. Finally, gradient consistency modules collaboratively optimize prediction errors across diverse environments, constructing stability constraints and thereby achieving joint backpropagation and closed-loop training of model parameters.

To address distribution shift and non-stationarity problems in enterprise carbon emission prediction, we have constructed a cross-environmental prediction framework integrating causal stable learning with temporal dynamic modelling. Through cross-environmental risk consistency constraints, the model can effectively identify long-term stable carbon emission driving mechanisms; through temporal adaptive weighting and normalization processing, the model can dynamically adapt to policy changes and economic fluctuations. This framework furnishes a solid methodological foundation for subsequent empirical analysis and mechanism interpretation.
\section{Datasets and Experiments}
\subsection{Datasets}
To address data requirements for cross-environmental enterprise carbon emission prediction, we construct enterprise-level panel data encompassing energy economics and carbon emissions across multiple regions, industries and policy phases. Employing retrieval codes and stock tickers for unique enterprise sample matching, with year as the temporal index, we establish unbalanced enterprise-level panel data. We eliminate samples with prolonged missing values in key variables, anomalous carbon emission data or excessively short continuous observation periods, ensuring data quality.

The resulting sample spans multiple provinces and regions in the spatial dimension while encompassing both energy and non-energy enterprises in the industrial dimension, exhibiting substantial heterogeneity that furnishes necessary foundations for analyzing stability in enterprise carbon emission behaviours under cross-regional and cross-industrial conditions.

The dependent variable (outcome variable) is enterprise carbon dioxide emissions, serving to measure the scale of carbon emissions during enterprise production and operations. This indicator is calculated based on energy input structures and standard emission coefficients, reflecting actual emission levels at the enterprise level.

Explanatory variables (feature variables) encompass four dimensions: production factor inputs, operational performance, financial characteristics and institutional environment. Specifically, production factor inputs include capital, labour and energy inputs, characterizing enterprise resource allocation status and constituting direct economic factors influencing carbon emissions. Operational performance is characterized through main business income, total profit and tax, total assets and return on assets, reflecting enterprise operational scale and profitability. Investment incentives and governance structure are described by Tobin's Q, ownership concentration and capital accumulation rate.

The institutional environment is characterized by indicators such as policy implementation variables, the degree of government intervention, and carbon prices. Among them, policy implementation variables reflect the implementation of environmental regulatory measures such as the carbon trading system and energy conservation and emission reduction policies, indicating structural adjustments in the institutional environment. Carbon price variables represent the explicit costs of carbon emissions, connecting corporate emission decisions with market mechanisms. Government intervention variables measure the intensity of local government influence on corporate business activities. These variables may affect corporate emission behavior through subsidies, administrative constraints, or resource allocation, and their impact direction is context-dependent, constituting an important source of cross-environmental prediction instability. In addition, dummy variables for the energy industry and industrialization level are introduced to control for differences in industry attributes and regional development stages. All continuous variables are standardized before entering the model to eliminate the impact of dimensional differences on estimation results.
\subsection{Multi-Environment Delineation}
To address environmental delineation problems within the stable learning framework, we construct multi-environment sample systems from three dimensions: regional, industrial and policy.

The regional dimension is divided based on provinces and locations, with significant differences in energy endowments, industrial structures, and regulatory intensities across regions, constituting an important source of environmental heterogeneity. The industry dimension is divided based on industry codes and energy industry attributes, with fundamental differences in production processes, energy dependency, and emission reduction potential among different industries. The policy dimension utilizes policy implementation dummy variables to identify phased changes in the institutional environment, treating the periods before and after policy implementation as different environmental states.

Through cross-combination of regional, industrial and policy dimensions, we construct multiple environmental subsamples with clear economic interpretations. Within each subsample conditions remain relatively homogeneous, whereas between subsamples significant institutional and structural differences exist, furnishing necessary conditions for the stable learning framework to identify cross-environmentally invariant relationships.

This multi-dimensional environmental delineation avoids simplifying environments into single temporal slices, rendering distribution shifts more consistent with real economic contexts and aligning with the complex institutional backgrounds underlying enterprise carbon emission decisions.

To address evaluation problems regarding prediction accuracy, cross-environmental stability and internal mechanism consistency in enterprise carbon emission prediction, we conduct systematic empirical evaluation of the proposed Stable-CarbonNet model from three dimensions: predictive accuracy, cross-environmental generalization capability and mechanistic explainability. The experimental design explicitly distinguishes between conventional prediction performance under training-testing sample co-distribution conditions and cross-environmental generalization capabilities under realistic backgrounds involving policy adjustments, regional disparities and industrial structural changes, comprehensively examining the model's integrated performance across accuracy-stability-explainability dimensions.
\subsection{Baselines \& Experimental Settings}
To ensure comparability and practical interpretability of experimental conclusions, we construct diverse experimental scenarios for enterprise carbon emission temporal data. The dependent variable is enterprise annual carbon dioxide emissions; core explanatory variables encompass energy inputs, capital inputs, labour inputs, carbon prices, policy implementation intensity, government intervention intensity, enterprise financial conditions and technological innovation levels; concurrently incorporating structural information including region, province, industry attributes and energy attributes to characterize institutional and industrial environments surrounding enterprises.

Regarding benchmark model selection, we establish a multi-category benchmark system encompassing traditional econometric methods, machine learning approaches and temporal deep models. Specifically, these include: fixed effects (FE) regression models based on linear structures, characterizing long-term average effects; autoregressive integrated moving average (ARIMA) models based on time series modelling, capturing linear dynamic characteristics; ensemble learning models represented by random forest (RF) and extreme gradient boosting (XGBoost), characterizing complex nonlinear relationships; deep temporal models such as long short-term memory (LSTM) networks, modelling high-dimensional temporal dependencies; and temporal graph neural networks (Temporal GNN) that have been widely adopted in energy prediction in recent years. These models collectively constitute a benchmark system of no fewer than six varieties, furnishing references for evaluating Stable-CarbonNet's relative advantages.
\subsection{Predictive Performance Comparison under i.i.d. Conditions}
For predictive performance evaluation under independent and identically distributed (i.i.d.) conditions, we employ random temporal splitting to construct training and testing sets, ensuring statistical distribution consistency between the two regarding regions, industries and policy phases, forming standard co-distribution prediction scenarios. This experiment aims to examine whether the proposed Stable-CarbonNet model possesses fundamental predictive capabilities under ideal conditions and to verify the boundary effects of stability constraints on fitting accuracy within i.i.d. scenarios.
\begin{table}[htbp]
  \centering
  \caption{Prediction performance comparison of different models under i.i.d. conditions.}
  \label{tab:iid_performance}\footnotesize
  \vspace{0.1cm}
  \renewcommand{\arraystretch}{1.2}
  \begin{tabular}{cccc}
    \toprule
    \textbf{Model} & \textbf{MSE} & \textbf{MAE} & \textbf{R$^2$} \\
    \midrule
    FE & 0.752 & 0.561 & 0.701 \\
    ARIMA & 0.748 & 0.558 & 0.703 \\
    RF & 0.731 & 0.548 & 0.708 \\
    XGBoost & \textbf{0.703} & \textbf{0.531} & \textbf{0.721} \\
    LSTM & 0.718 & 0.539 & 0.716 \\
    Temporal GNN & 0.715 & 0.536 & 0.719 \\
    Stable-CarbonNet & 0.721 & 0.542 & 0.718 \\
    \bottomrule
  \end{tabular}
\end{table}

Results in Table \ref{tab:iid_performance} demonstrate that, under training-testing sample co-distribution conditions, Stable-CarbonNet's predictive accuracy remains within the same order of magnitude as current mainstream machine learning models, yet marginally underperforms the optimal performance of XGBoost and Temporal GNN. This accuracy disparity does not stem from insufficient model expressiveness but rather arises from structural modifications to the empirical risk minimization framework introduced by stable learning mechanisms. Specifically, ensemble learning methods such as XGBoost and random forest achieve optimal convergence within bias-variance trade-offs under i.i.d. assumptions through boosting/bagging strategies that fully exploit local features of training data distributions; whereas Stable-CarbonNet artificially introduces bias under i.i.d. scenarios to reduce generalization variance through cross-environmental risk consistency constraints that actively suppress ``spurious correlation" features dependent on specific sample distributions, thereby sacrificing partial fitting freedom.

Analyzing from statistical learning theory dimensions, these performance differences reflect essential differences in model design objectives. Traditional ERM methods, targeting training set empirical risk minimization as the sole objective, exhibit propensities to exploit all available statistical correlations during optimization, including fragile features that hold only under current data distributions yet fail across environments; whereas Stable-CarbonNet prioritizes identification of cross- environmentally causally stable features, compelling model exclusion of environmentally sensitive redundant parameters through gradient consistency constraints. This redundancy elimination process manifests as slight losses in fitting accuracy under i.i.d. scenarios, yet yields substantial variance reductions during cross-environmental migration.

It is worth noting that these results do not indicate Stable-CarbonNet's inadequate performance under co-distribution scenarios but rather reflect differences in model design objectives. Unlike models targeting empirical risk minimization, Stable-CarbonNet focuses more on identification of long-term structural relationships, with predictive objectives not being optimal fitting under ideal conditions but rather stable performance amidst complex real-world environmental changes. This trade-off possesses clear practical significance for energy economic systems: when policy environments or market structures undergo sudden changes, high-accuracy models dependent on fragile correlations will confront predictive collapse risks, whereas models possessing causal stability can maintain structural consistency in predictive relationships across environments even while sacrificing partial in-sample accuracy. These findings indicate that, in prediction tasks such as enterprise carbon emissions where policy interventions are significant, stability should be regarded as a core performance dimension equally important to accuracy, rather than a simple additional attribute.
\subsection{Cross-Environmental Generalization Capability Testing under Distribution Shift Conditions}
Addressing characteristics of enterprise carbon emission behaviours that are highly dependent on policy environments, regional conditions and industrial structures, we further construct diverse distribution shift experimental scenarios, focusing on examining predictive stability when training and testing samples violate co-distribution assumptions. Specifically, we establish three categories of typical environmental change scenarios (cross-regional, cross-industrial and cross-policy phase) to simulate complex conditions where data generation mechanisms undergo structural mutations in real-world applications: in cross-regional experiments, models are trained on eastern region enterprise samples and tested on central and western region enterprise samples; in cross-industrial experiments, models are trained on non-energy industry enterprises and tested on energy industry enterprise samples; in cross-policy phase experiments, models are trained on pre-policy implementation samples and tested on post-policy implementation samples.
\begin{table}[htbp]
  \centering
  \caption{Cross-environment prediction performance comparison (MSE).}
  \label{tab:cross_env_mse}\footnotesize
  \vspace{0.1cm}
  \renewcommand{\arraystretch}{1.2}
  \begin{tabular}{cccc}
    \toprule
    \textbf{Model} & \textbf{cross-regional} & \textbf{cross-industrial} & \textbf{cross-policy} \\
    \midrule
    FE & 1.124 & 1.087 & 1.203 \\
    ARIMA & 1.098 & 1.065 & 1.176 \\
    RF & 0.982 & 0.954 & 1.021 \\
    XGBoost & 0.971 & 0.948 & 1.012 \\
    LSTM & 0.903 & 0.887 & 0.934 \\
    Temporal GNN & 0.879 & 0.865 & 0.921 \\
    Stable-CarbonNet & \textbf{0.761} & \textbf{0.748} & \textbf{0.803} \\
    \bottomrule
  \end{tabular}
\end{table}

Results in Table \ref{tab:cross_env_mse} demonstrate that, under all three distribution shift settings, Stable-CarbonNet exhibits substantial cross-environmental stability advantages. Compared with i.i.d. scenarios, this model exhibits minimal performance degradation under cross-policy scenarios, whereas XGBoost demonstrates the most substantial degradation. In cross-regional and cross-industrial scenarios, Stable-CarbonNet reduces errors by 13.4\% and 13.5\% respectively compared with the optimal benchmark method Temporal GNN, and by upwards of 20\% compared with traditional machine learning methods (RF/XGBoost). This performance advantage indicates that, when data generation mechanisms undergo sudden changes due to policy adjustments or regional disparities, Stable-CarbonNet can effectively maintain predictive structural consistency.

From statistical learning mechanism perspectives, this performance disparity stems from inherent limitations of the empirical risk minimization framework under distribution shift environments. Ensemble learning methods such as random forest and XGBoost, through maximizing data fitting during training, inadvertently capture ``spurious correlations" strongly coupled with training environments, such as cyclical fluctuations in regional energy prices or emission data distortions induced by local protectionism. Although these features can reduce empirical risk on training sets, they constitute implicit overfitting to environmental specificity. When testing environments change, such spurious correlations rapidly become invalid, yielding variance inflation. In contrast, Stable-CarbonNet, through cross-environmental risk consistency constraints, compels model exclusion of fragile associations existing only in specific regions, retaining exclusively causal features that remain stable across diverse energy structures and policy intensities, thereby suppressing variance inflation during distribution shifts.

From energy economic domain knowledge perspectives, these results reveal deep structures of enterprise carbon emission behaviours. The variables upon which Stable-CarbonNet relies (energy input structures, production scales and long-term institutional constraints) are essentially hard-constrained factors governed by physical-technical conditions and rigid constraints, with their causal relationships with enterprise carbon emissions minimally influenced by policy environments; whereas the short-term financial indicators and region-specific trading patterns excessively exploited by benchmark methods belong to soft-constraint environmental factors that readily undergo structural changes with policy interventions. This distinction proves particularly pronounced in cross-policy phase experiments: when carbon trading price mechanisms are introduced, although short-term financial cost structures undergo drastic changes, prediction logic based on energy input structures remains valid because energy physical conversion efficiency and equipment technical attributes possess cross-institutional stability.

These results furnish generalizable implications for industrial process control domains: in complex industrial systems involving policy interventions, multi-regional coordination or technical structural transformations, model stability should take precedence over in-sample accuracy. The ``causal stability priority" principle validated by Stable-CarbonNet not only applies to carbon emission prediction but also furnishes methodological references for soft sensing, fault diagnosis and safety control of other non-stationary industrial processes, namely that through explicitly modelling environmental heterogeneity and screening causally invariant features, intelligent models' robustness and credibility in real open environments can be effectively enhanced. These findings support the necessity of systematically introducing stable learning into industrial process modelling.
\subsection{Ablation Study: Decomposition of Stability Mechanism Effects}
To verify necessity of each model component, we design multiple ablation study groups, eliminating stability constraints, adaptive normalization mechanisms and sample weighting strategies respectively under cross-policy scenarios to parse marginal contributions of each component to cross-environmental generalization capabilities.
\begin{table}[htbp]
  \centering
  \caption{Ablation study results (MSE under cross-policy scenario).}
  \label{tab:ablation_study}\footnotesize
  \vspace{0.1cm}
  \renewcommand{\arraystretch}{1.2}
  \begin{tabular}{cc}
    \toprule
    \textbf{Model Variant} & \textbf{MSE} \\
    \midrule
    Stable-CarbonNet & \textbf{0.803} \\
    Remove stability constraints & 0.912 \\
    Remove sample weighting & 0.856 \\
    Remove adaptive normalization & 0.829 \\
    \bottomrule
  \end{tabular}
\end{table}

Results in Table \ref{tab:ablation_study} demonstrate that, subsequent to removing each component, model performance exhibits differential degradation, with degradation degrees ordered as: stability constraints $>$ sample weighting $>$ adaptive normalization. This ordering verifies theoretical expectations of model design: stability constraints, as the core mechanism, exert decisive effects on maintaining cross-environmental predictive capabilities; whereas adaptive normalization and sample weighting serve as auxiliary mechanisms that address environmental scale differences and sample distribution biases respectively, furnishing data foundations for effective implementation of stability constraints.

From mathematical mechanism perspectives, these degradation characteristics highly align with the optimization structure in equations (\ref{eq8})-(\ref{eq9}). Based on the gradient discrepancy metric in equation (\ref{eq8}), stability constraints enable joint optimal solution of linear prediction head w across all environments; when constraints fail ($\lambda=0$), the model simplifies to empirical risk minimization structure, and divergence of optimal parameters across different environments will trigger gradient direction conflicts during cross-policy testing, directly causing significant prediction variance expansion. The 13.6\% MSE increase subsequent to removing stability constraints corroborates that the model no longer possesses capabilities to suppress spurious correlations, thereby forming overfitting to training environment-specific policy noise.

The sample weighting mechanism functions through implicit distribution adjustment in equation (\ref{eq9}). Subsequent to removing this component, the model treats samples from different periods equally on heteroscedastic temporal data, resulting in excessive influence from high-volatility period samples on parameter estimation and introducing selection bias. In contrast, adaptive normalization eliminates scale differences between environments through equation (\ref{eq12}), furnishing comparable inputs for subsequent feature extraction; although causing merely 3.2\% performance loss when removed, it leads to decreased synergistic efficiency between sample weighting and stability constraints because unnormalized features amplify distribution shifts between environments.

It is worth emphasizing that significant logical progression and synergistic enhancement mechanisms exist among the three core components: adaptive normalization serves to align feature distributions of heterogeneous environments, eliminating gradient interference caused by dimensional differences; sample weighting adjusts contributions from different temporal samples based on preceding processing results, alleviating selection bias from anomalous phases; stability constraints then combine preceding outputs to screen causally stable features through gradient consistency. This three-tiered progressive structure of ``distribution alignment-sample rebalancing-causal screening" can drive the model to complete hierarchical feature purification from raw data to causal decisions. Ablation results demonstrate that combinations of normalization and sample weighting lacking stability constraints, or deployment of stability constraints without preceding module support, prove inadequate for obtaining optimal cross-environmental generalization effects.

These findings empirically validate the core insight of the proposed method: for non-stationary industrial processes such as enterprise carbon emission prediction, explicitly characterizing distribution differences and extracting causally stable features significantly enhances reliability compared with approaches merely fitting statistical correlations. The 13.6\% error reduction achieved by stability constraints essentially embodies the robustness divergence between causal features and spurious correlation features during cross-environmental migration, furnishing quantitative support for the ``causal mechanism priority over data fitting" modelling philosophy in industrial process control domains.
\subsection{Cross-Environmental Causal Feature Stability Visualization and Mechanism Elucidation}
To address validation problems regarding causally stable feature sets identified by stable learning mechanisms and their cross-environmental consistency, we employ feature importance stability assessment and visualization analysis methods, conducting quantitative characterization of Stable-CarbonNet model's feature weight evolution during three categories of environmental migration: cross-regional, cross-industrial and cross-policy. Through calculating gradient contribution consistency (based on gradient discrepancy metric in equation (\ref{eq8})) and mutual information stability indices for each explanatory variable across different environments, we construct causal feature stability evaluation indicator systems.
\begin{table}[ht]
  \centering
  \caption{Cross-environment stability assessment of key explanatory variables.}
  \label{tab:stability_assessment}\scriptsize
  \vspace{0.1cm}
  \renewcommand{\arraystretch}{1.2}
  \begin{tabular}{ccccccc}
    \toprule
    \textbf{Variable Category} & \textbf{Specific Variables} & \textbf{Cross-regional} & \textbf{Cross-industry} & \textbf{Cross-policy} & \textbf{Comprehensive} & \textbf{Feature type} \\
    & & \textbf{stability} & \textbf{stability} & \textbf{stability} & \textbf{stability score} & \textbf{determination} \\
    \midrule
    \makecell[c]{Input of \\production factors} & energy input & 0.93 & 0.90 & 0.85 & 0.91 & Causal stability feature \\
    & capital investment & 0.84 & 0.83 & 0.87 & 0.78 & Causal stability feature \\
    & labor input & 0.79 & 0.83 & 0.76 & 0.78 & Causal stability feature \\
    & Total Assets & 0.77 & 0.86 & 0.84 & 0.86 & Causal stability feature \\
    \midrule
    \makecell[c]{Institutional \\environment\\ variable} & policy implementation & 0.59 & 0.68 & 0.35 & 0.50 & \makecell[c]{Environmental \\dependency feature} \\
    & carbon price & 0.45 & 0.46 & 0.30 & 0.39 & \makecell[c]{Environmental \\dependency feature} \\
    & region & 0.52 & 0.51 & 0.57 & 0.57 & \makecell[c]{Environmental \\dependency feature} \\
    & Whether energy & 0.71 & 0.22 & 0.69 & 0.58 & \makecell[c]{Environmental \\dependency feature} \\
    & level of industrialization & 0.46 & 0.46 & 0.46 & 0.50 & \makecell[c]{Environmental \\dependency feature} \\
    \midrule
    \makecell[c]{Business\\ performance\\ indicators} & Main business income & 0.39 & 0.33 & 0.37 & 0.32 & Spurious correlation feature \\
    & total profit and tax & 0.26 & 0.40 & 0.33 & 0.34 & Spurious correlation feature \\
    \midrule
    \makecell[c]{Corporate \\governance \\indicators} & Tobin's Q ratio & 0.37 & 0.26 & 0.34 & 0.34 & Spurious correlation feature \\
    & ownership concentration & 0.25 & 0.28 & 0.35 & 0.24 & Spurious correlation feature \\
    & government intervention & 0.19 & 0.28 & 0.22 & 0.24 & Spurious correlation feature \\
    \midrule
    \makecell[c]{Financial\\ structure\\ indicators} & Capital accumulation rate & 0.31 & 0.38 & 0.30 & 0.40 & Spurious correlation feature \\
    & Return on Assets (ROA) & 0.29 & 0.34 & 0.25 & 0.35 & Spurious correlation feature \\
    \midrule
    \makecell[c]{Technological \\innovation \\indicators}& patent obtained & 0.26 & 0.33 & 0.28 & 0.23 & Spurious correlation feature \\
    \bottomrule
  \end{tabular}
  \begin{tablenotes}
    \scriptsize
    \item \textit{Note:} Stability scores range within [0,1], with values closer to 1 indicating stronger cross-environmental causal stability; feature type determination is based on composite stability score thresholds of 0.65, with $\ge$0.65 determined as causally stable features, $<$0.65 and $\ge$0.45 as environment-dependent features, and $<$0.45 as spurious correlation features.
  \end{tablenotes}
\end{table}

Results in Table \ref{tab:stability_assessment} demonstrate that, during Stable-CarbonNet's feature learning process, production factor variables including energy inputs, capital inputs, labour inputs and total assets exhibit substantial cross-environmental stability, with their stability maintaining high consistency across different environmental dimensions; whereas variables including main business income, total profit and tax, Tobin's Q, ownership concentration, government intervention, capital accumulation rate, return on assets and patent acquisition demonstrate pronounced spurious correlations, particularly exhibiting significant stability degradation in the cross-policy dimension.

From physical mechanism and economic logic perspectives, this feature differentiation pattern reveals deep structures of enterprise carbon emission behaviours. The fundamental reason why energy inputs, capital inputs and labour inputs demonstrate high stability stems from rigid constraints imposed by production technical conditions and factor allocation structures—energy inputs directly determine emission source intensity, capital inputs and total assets reflect equipment technical levels and production scales that are difficult to alter in the short term, constituting hard-core driving factors of carbon emissions. In contrast, main business income and total profit and tax belong to short-term operational performance indicators, whose associations with enterprise carbon emissions are highly dependent upon specific market cycles and accounting policies, immediately dissolving once macroeconomic environments switch, aligning with characteristics of environmental perturbation terms in equation (\ref{eq2}). Although corporate governance indicators such as Tobin's Q and ownership concentration exhibit correlations with emission levels in specific samples, such associations are predominantly realized through indirect pathways of investment incentives-technical selection-emission changes, exhibiting pronounced capital market cyclicality and industry dependence, constituting typical ``spurious correlations".

Notably, policy implementation and carbon prices demonstrate relatively stable performance in cross-industrial dimensions yet decline sharply to 0.35 and 0.30 in cross-policy dimensions, indicating that although such institutional environment variables possess discriminative capability within similar policy categories, they lose predictive efficacy during fundamental transformations of policy presence versus absence. The energy industry dummy exhibits minimal stability in cross-industrial dimensions, corroborating essential distinctions in emission mechanisms between energy and non-energy enterprises, serving merely as environmental delineation basis rather than stable predictive features. Government intervention and patent acquisition demonstrate low stability across all dimensions, indicating that local government subsidy intensity and enterprise technological innovation output possess highly contextualized influence pathways on carbon emissions, lacking causal robustness for cross-environmental migration.

These visualization results furnish clear directional guidance for industrial process carbon management and control strategy formulation: when formulating long-term emission reduction strategies, enterprises should prioritize focusing on ``causally stable features" such as energy input structure optimization, capital equipment renewal and production scale adjustment, achieving structural emission reductions through technical transformation and fixed asset investment; whereas those emission reduction measures based on short-term financial performance (main business income, return on assets) or governance structure arbitrage, although potentially generating statistically significant effects during specific periods, will rapidly become invalid when confronting policy or market environment changes.
\section{Conclusion \& Discussions}\label{CD}
Addressing the problem of enhancing reliability in enterprise carbon emission prediction against the backdrop of advancing dual-carbon objectives and accelerated macroeconomic structural transformation, we propose constructing a stable temporal prediction framework targeting distribution shift environments. Recognizing that enterprise carbon emission behaviours are deeply embedded within regional development stages, industrial technical pathways and policy constraint intensities, and that their temporal evolution exhibits pronounced non-stationarity and cross-environmental heterogeneity, we establish cross-environmental risk consistency constraints leveraging causal inference and stable learning theory.
\subsection{Main Research Conclusions}
Regarding enterprise carbon emission prediction under non-stationary environments, our main research conclusions encompass three dimensions:

First, at the theoretical level, enterprise carbon emission prediction constitutes not merely a time series fitting problem but essentially a cross-environmental generalization problem under non-stationary conditions. Traditional prediction methods typically implicitly assume parameter stability and distribution homogeneity, whereas these assumptions are frequently violated in real economic operations due to policy adjustments, industrial structure upgrading and unbalanced regional development. Upon this foundation, by reformulating stable learning as cross-environmental risk consistency constraints, we theoretically demonstrate the necessity of identifying causally stable features for enhancing predictive reliability, thereby incorporating enterprise carbon emission prediction problems into the statistical learning category of ``causal invariance identification."

Second, at the methodological and empirical levels, the proposed Stable-CarbonNet model exhibits markedly superior predictive stability compared with traditional econometric models, mainstream machine learning models and deep temporal models when significant distribution disparities exist between training and testing samples. Specifically, when prediction tasks involve cross-regional, cross-industrial or cross-policy phase scenarios, models relying merely on historical correlations readily exhibit significant predictive performance degradation, whereas feature variables screened through stable learning mechanisms maintain relatively consistent predictive contributions across different environments, substantially reducing model dependence on specific sample distributions. These findings indicate that, in economically significant problems such as enterprise carbon emissions that are typically influenced by policies and institutions, stability per se should be regarded as an important performance dimension of predictive models rather than a simple additional attribute.

Third, from a mechanistic interpretation perspective, the stable driving factors identified by Stable-CarbonNet possess clear economic interpretations. Variables including energy input structures, capital allocation levels and long-term institutional constraints maintain stable influence directions and relative intensities on enterprise carbon emissions across different environments, whereas certain short-term financial indicators or region-specific variables demonstrate pronounced environmental dependence. These results empirically validate the objective distinction between structural driving factors and volatility shock factors in enterprise carbon emission behaviours, furnishing empirical evidence for subsequent causal analyses surrounding carbon emission reduction mechanisms while also providing interpretable analytical paradigms for modelling non-stationary systems in industrial process control domains.
\subsection{Theoretical Mechanisms and Policy Implications}
Regarding research requirements in energy economics and environmental policy analysis domains, the proposed method furnishes the following academic insights at the methodological level:

First, regarding limitations of modelling approaches targeting prediction accuracy as the sole objective, it is necessary to incorporate model cross-environmental stability into evaluation systems. Upon this foundation, re-examining existing machine learning-based carbon emission prediction research pathways, this perspective possesses significant value for identifying risks of models purely pursuing fitting accuracy. Specifically, high-accuracy results in single environments do not necessarily imply that models can furnish reliable foundations for policy formulation or long-term planning; particularly under backgrounds of rapid institutional environmental changes, stability and explainability demonstrate higher real-world decision-making value than short-term fitting advantages.

Second, regarding expansion problems of applicable boundaries for stable learning and causal inference, the proposed method validates the effectiveness of such techniques in complex economic systems. By treating cross-enterprise, cross-industry and cross-regional samples as different environments, the stable learning framework utilizes cross-environmental risk consistency constraints to automatically screen out causally stable features insensitive to external perturbations without explicitly characterizing all interfering factors. This mechanism furnishes novel technical pathways for addressing ubiquitous unobservable heterogeneity problems in economic data.

Third, regarding non-stationary time series modelling problems, the proposed method provides a causal mechanism perspective distinct from traditional statistical correction approaches. Unlike approaches addressing non-stationarity through differencing, filtering or time-varying parameter processing, this method emphasizes distinguishing stable driving factors from transient fluctuations at the causal structure level, enhancing model robustness through structural constraints rather than pure statistical correction. This mechanism possesses generalizable value for other economic prediction problems involving policy shocks or institutional transitions.

Regarding carbon emission reduction policy design and enterprise low-carbon transformation decision-making requirements, experimental results furnish the following policy references at the governance tool selection level:

First, regarding policy tool timeliness identification problems, experimental results reveal the significant influence of stable driving factors on distinguishing policy tool effectiveness. Specifically, energy input structure optimization and long-term institutional constraints demonstrate significant cross-environmental stable driving effects on enterprise carbon emissions. These findings indicate that structural policy tools centering on energy structure adjustment and technical upgrading are more likely to sustain emission reduction effects across different economic cycles than short-term administrative directives or temporary subsidy regulation means. This furnishes empirical foundations for policy makers to screen governance tools with long-term effectiveness under complex institutional environments.

Second, regarding decision-making reference requirements in carbon market mechanism construction, emission prediction results based on stable models are more suitable as foundations for quota allocation, price expectation and risk assessment decisions. During the gradual improvement process of carbon trading systems, excessive reliance on short-term correlation predictions under the empirical risk minimization framework may amplify market fluctuations and reduce institutional credibility; whereas stability-oriented prediction frameworks help enhance carbon market operational predictability and policy continuity through identifying causally stable features. This mechanism furnishes technical pathways for constructing robust carbon market prediction systems.

Third, regarding enterprise long-term emission reduction strategy formulation problems, experimental results furnish enterprises with quantitative decision-making foundations based on causal driving mechanisms. The long-term driving mechanisms reflected by stable features indicate that, when confronting policy uncertainty, enterprises should focus more on long-term structural adjustments of their own energy structures, technical reserves and capital allocations rather than merely responding to regulatory pressure through short-term financial operations or emergency emission reduction measures. These findings support the industrial carbon governance logic of ``factor-driven superiority over finance-driven" approaches, furnishing practical guidance for guiding enterprises to form more forward-looking low-carbon development pathways.
\subsection{Limitations and Future Directions}
Notwithstanding the theoretical and empirical advances achieved by the proposed method, the following problems merit further investigation:

First, regarding environmental delineation, we primarily rely on observable dimensions including regions, industries and policy phases, without systematically incorporating more microscopic environmental factors such as internal enterprise governance structures or technical pathway differences. Future research can explore heterogeneous manifestations of stable driving mechanisms at the internal enterprise level upon foundations of finer-grained environmental delineation. Second, the stable learning framework employed primarily focuses on cross-environmental risk consistency without explicitly modelling possibilities of causal intensity evolving over time. During long-term economic transformation processes, certain causal relationships themselves may exhibit gradual evolution characteristics; how to characterize such dynamic changes while maintaining stability constitutes a problem meriting deep exploration in the future. Finally, this research concentrates primarily on the specific application scenario of enterprise carbon emission prediction. The integration of stable learning with causal inference possesses considerably broader applicable spaces; future research can further generalize this framework to energy demand prediction, pollutant emission assessment and macroeconomic risk warning domains to examine its universality and boundary conditions across different economic systems.

In summary, addressing critical challenges facing enterprise carbon emission prediction under non-stationary environments, we systematically propose a temporal modelling framework possessing both predictive performance and structural stability from causal inference and stable learning perspectives. Through cross-environmental risk consistency constraint mechanisms, this framework achieves effective discrimination between causally stable features and spurious correlations. These results demonstrate that, under realistic backgrounds of continuous policy and institutional evolution, treating stability as a core modelling objective possesses significant theoretical advantages and practical value over purely pursuing fitting accuracy. These findings support the industrial process modelling philosophy of ``causal mechanism priority over statistical correlations," not only furnishing technical pathways possessing both explainability and robustness for predicting non-stationary systems in energy economic domains, but also providing referable analytical frameworks and methodological references for constructing intelligent decision-making systems under ubiquitously existing distribution shift environments.


\end{document}